\documentclass{article}
\usepackage{ijcai20}

\usepackage{times}

\usepackage{soul}
\usepackage{url}
\usepackage[hidelinks]{hyperref}
\usepackage[utf8]{inputenc}
\usepackage[small]{caption}
\usepackage{graphicx}
\usepackage{amsmath}
\usepackage{amsfonts}
\usepackage{booktabs}
\newcommand\ourgan[0]{{PCTGAN}}
\usepackage{lipsum}

\usepackage{algorithm,algorithmic}
\newcommand\blfootnote[1]{%
  \begingroup
  \renewcommand\thefootnote{}\footnote{#1}%
  \addtocounter{footnote}{-1}%
  \endgroup
}

\usepackage{comment}

\title{Physical Context and Timing Aware Sequence Generating GANs}

\author{
Hayato Futase$^1$\footnote{Contact Author. The author is now at Osaka University.}\and
Tomoki Tsujimura$^1$\and
Tetsuya Kajimoto$^2$\and
Hajime Kawarazaki$^2$\and
Toshiyuki Suzuki$^2$\and
Makoto Miwa$^1$\and
Yutaka Sasaki$^1$\and
\affiliations
$^1$Toyota Technological Institute\\
$^2$Toyota Motor Corporation\\
\emails
h-futase@ei.sanken.osaka-u.ac.jp$^{*}$
}

\begin{document}

\maketitle

\begin{abstract}
\blfootnote{Preprint.}
Generative Adversarial Networks (GANs) have shown remarkable successes in generating realistic images and interpolating changes between images. 
Existing models, however, do not take into account physical contexts behind images in generating the images, which may cause unrealistic changes. Furthermore, it is difficult to generate the changes at a specific timing and they often do not match with actual changes.
This paper proposes a novel GAN, named Physical Context and Timing aware sequence generating GANs (\ourgan{}), that generates an image in a sequence at a specific timing between two images with considering physical contexts behind them. 
Our method consists of three components: an encoder, a generator, and a discriminator. The encoder estimates latent vectors from the beginning and ending images, their timings, and a target timing. The generator generates images and the physical contexts at the beginning, ending, and target timing from the corresponding latent vectors. The discriminator discriminates whether the generated images and contexts are real or not. In the experiments, \ourgan{} is applied to a data set of sequential changes of shapes in die forging processes. We show that both timing and physical contexts are effective in generating sequential images. 
\end{abstract}

\maketitle

\section{Introduction}

\begin{figure}[t!]
\centering
\includegraphics[width=\linewidth]{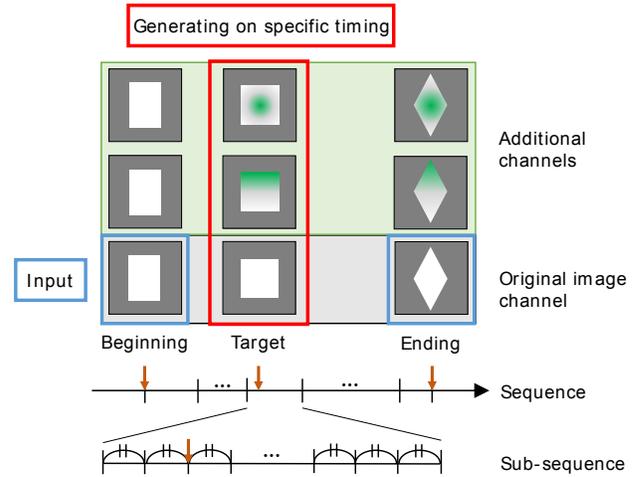}
\caption{Sequence of images with physical contexts and timing.}
\label{fig:phyandtime}
\end{figure}

Generative Adversarial Networks (GAN)~\cite{NIPS2014_5423} has been successful in image generation tasks. GAN is composed of a generator and a discriminator and they are trained so that the discriminator can discriminate real images from the images generated by the generator. As a result, the generator learns to generate an image that is close to a real image. 
There are a few studies that use physical constraints in generating images. Generation of images with enforcing constraints on co-variance of the training data is proposed to generate images of Rayleigh-B\'enard convection~\cite{FlowImage}. The loss with explicit physical constraints is also proposed~\cite{physicalconstrain}.

In generating images, translation between images is also performed with GAN~\cite{pix2pix,CycleGAN,StarGAN}. Furthermore, translation of images into other domain information is proposed such as translating RGB images into images with depth information~\cite{depthfromRGBvideo} and translating text to images~\cite{Text2Image}.  Furthermore, there are some studies in video interpolation. Most studies in video interpolation generate middle image among three adjacent frames for short-term videos~\cite{superslomo,choi2020cain}. Only two studies focus on long-term videos~\cite{FromHeretoThere,xu2018stochastic}, i.e., low fps or not using adjacent frames. One is generating a certain number of frames, and the other is generating images between two intermediate images in addition to beginning and ending images~\cite{xu2018stochastic}.

Conventional studies, however, do not consider their physical contexts into account in generating images. Furthermore, the timing of the intermediate image to be generated in a sequence is not considered, although the sequence could be modeled more accurately by enabling the generation of an image at an arbitrary timing in a sequence.
Figure~\ref{fig:phyandtime} illustrates the sequence of changes in die forging data. Forging is a process in which the target shape is finally obtained by repeatedly placing metal in the die and deforming it. 
It is necessary to evaluate the metal state at an arbitrary timing, and the data have physical contexts such as stress and strain, which are shown in additional channels of the Figure~\ref{fig:phyandtime}. 

Furthermore, no GANs consider the timing of the intermediate image to be generated in a sequence, although the sequence could be modeled more accurately by enabling the generation of an image at an arbitrary timing in a sequence. Most studies in video interpolation generate middle image among three adjacent frames for short-term videos~\cite{superslomo,choi2020cain}. Only two studies focus on long-term videos~\cite{FromHeretoThere,xu2018stochastic}, i.e., low fps or not using adjacent frames. One is generating a certain number of frames, and the other is generating images between two intermediate images in addition to beginning and ending images~\cite{xu2018stochastic}, but they do not consider the timing. 


We propose a novel model, namely Physical Context and Timing aware sequence generating GANs (\ourgan{}), that generates an image at a specific timing between two images considering physical contexts behind them. Our target task is different from existing interpolation tasks in two ways.
First, \ourgan{} considers the timing of an image in a sub-sequence of a sequence. The sequence is divided into several sub-sequences so that the model can deal with discontinuous and different changes.  
Second, \ourgan{} deals with an original image channel as well as additional channels that correspond to physical contexts, such as speed and force. 
We assume that such timing and contexts are available in training. 
Unlike the original GAN, \ourgan{} consists of not only a generator and a discriminator but also an encoder that specifies the interpolated images and the target timing. 
Specifically, the encoder generates three latent vectors from two image channels that represent the beginning and ending of the sequence and the timing labels that specifies the beginning, ending, and target timings. The generator then generates images with all the channels including additional channels at beginning, ending, and the target timing from the latent vector. The discriminator discriminates the real or generated images into real or not from several viewpoints of channels using a channel label and the timing labels. By training these networks in an end-to-end manner, the generator is trained to generate realistic images constrained by both physical contexts and timing. In prediction, the encoder and generator generate the image sequence between two given images by generating images corresponding to specific timing labels one by one.

We evaluate \ourgan{} using die forging data. \ourgan{} generates an image of an intermediate shape in this deformation process by using shape information as an image channel and strain and stress as additional channels. In the experiment, we confirmed that \ourgan{} can generate consistent and realistic sequential images by using the timing and physical contexts.

The contribution of this paper is summarized as follows:
\begin{itemize}
\item Generation of an intermediate image at an arbitrary timing in the sequence by introducing the timing-label
\item Generation of realistic images by using physical contexts as additional channels
\item Generation of consistent and realistic image sequences by modelling the generator to generate all channels at an arbitrary timing and the discriminator to discriminate images conditioning on both timing and physical contexts
\end{itemize}

\section{Related Work}
\subsection{GANs}

In image generation, GANs repeat the process that the generator generates an image while the discriminator distinguishes a real image from a generated image, so that the generator will generate images that are close to real images. However, the original GANs suffered from gradient vanishing and mode collapse. To alleviate such problems, several methods such as Wasserstein GAN (WGAN)~\cite{WGAN}, WGAN-GP~\cite{WGAN-GP}, and SNGAN~\cite{SNGAN}, have been proposed. WGAN uses the Wasserstein-1 distance between real images and generated images. WGAN-GP further minimizes the following loss function to solve the problem that the weights in the discriminator of WGAN are likely to be localized.
\begin{equation}\label{eq:WGAN-GP}
\begin{split}
\mathbb{E}_{\widetilde{x} \sim P_{g}}[D(\widetilde{x})] - \mathbb{E}_{x \sim P_{r}}[D(x)] + \\
\lambda \mathbb{E}_{\widehat{x} \sim P_{\widehat{x}}}[(\parallel{\nabla_{\widehat{x}}}D(\widehat{x})\parallel_{2}-1)^{2}],
\end{split}
\end{equation}
where $P_{g}$, $P_{r}$, and $P_{\widehat{x}}$ are distributions of generated images, real images, and both images respectively. $D$ is a discriminator function and $\lambda$ is a gradient penalty coefficient. The third term, called gradient penalty, allows the distribution of all weights in the discriminator smooth and delivering gradients to further layers.
Furthermore, SNGAN proposed to normalize weights of each layer by the maximum singular value to make the discriminator K-Lipschitz continuous, which is shown to lead to the stability of learning~\cite{LS-GAN}. 

\subsection{Conditional GANs}

\begin{figure}[t!]
\centering
\includegraphics[width=\linewidth]{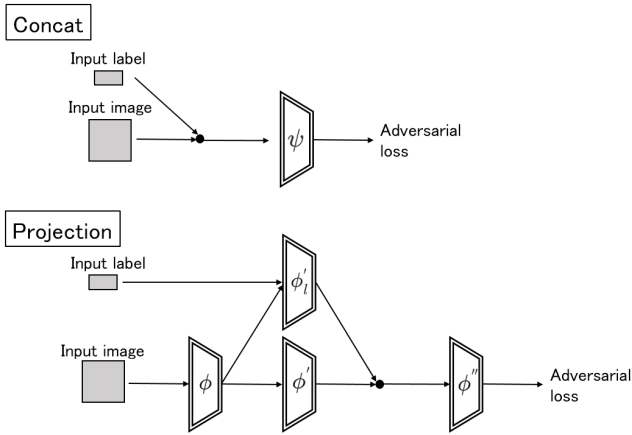}
\caption{Discriminator of cGAN models. $\psi$, $\phi$, $\phi'$, and $\phi''$ denote neural networks in the discriminator. }
\label{figure:concat}
\end{figure}

Conditional GANs (cGANs) are proposed to generate images in a desired class~\cite{conditionalGAN}. The original cGANs, a.k.a. \textit{Concat}, add an class label to input image and feeds them to the discriminator. This training is known to be unstable due to the sparse input. 
To resolve this problem, several improvements of cGANs have been proposed. Hidden Concat~\cite{Text2Image} concatenates the class label to a hidden layer in the discriminator. Projection Discriminator (\textit{Projection})~\cite{Projection_Disc} models label-dependent and label-independent representations and merge them. AC-GAN~\cite{AC-GAN} uses a subtask to predict class labels. Multi-Hinge Loss~\cite{Multi-Hinge_Loss} proposes an unified loss of the AC-GAN's loss and adversarial loss of GANs. 
We illustrate the discriminators of Concat and Projection in Figure~\ref{figure:concat}.

\subsection{Image Translation and Generation}

Image translation can be categorized into two types. One uses images in different domains without explicit correspondence~\cite{CycleGAN,StarGAN} and the other is based on the correspondence of images in different domains~\cite{pix2pix,pix2pixHD}. 
The former usually translates between RGB images, while the latter often translates RGB images into other domains like depth images~\cite{depthfromRGBvideo,depthfromRGBimg} and skeleton images~\cite{dance}. 

Recent methods propose generating images with physical information. Such methods include generating images of development of convection~\cite{FlowImage} and using the loss with physical constraints~\cite{physicalconstrain}. 
Furthermore, \cite{material-surface} estimated tactile properties from images by translating images into real values representing forces. 

\subsection{Video Interpolation}

Video interpolation methods can be mainly categorized into three types: optical flow-based, phase-based and pixels motion transformation methods.
Optical flow-based methods learn the model using optical flow among frames~\cite{Yu2013MultiLevelVF,voxelflow}. The estimation errors tend to be larger for longer sequences. Phase-based models~\cite{phasebase} treat each frame as a periodic wave with amplitude and phase and assume the differences in frames can be modeled with those in the wave. 
The modeling becomes difficult when they have to assume high-frequency wave for large motion videos. 
Pixels motion transformation methods directly estimate motions with deep learning without flows or phases. Such methods include methods generating an intermediate frame from two frames~\cite{niklaus2017video,LongTermVideo}, a method that considers reconstruction loss of multiple intermediate frames~\cite{superslomo}, and a method that generate intermediate frame sequences with gradually increasing the frame rates~\cite{FromHeretoThere}.

\section{\ourgan{}}

In this section, we explain \ourgan{} that generates sequential images between two images considering timing and additional physical context channels.
Unlike other sequence generation models, \ourgan{} directly generate an arbitrary intermediate image between two images instead of generating images in a specific temporal order.

Our task is an interpolation task in which the model needs to generate sequential images between two images, but our task is different from other interpolation tasks in two ways. 
First, we assume that a sequence is divided into a specific number of sub-sequences, so that we can model changes that are not continuous and contain several different changes.
Second, we assume that each image has their corresponding additional channels, which corresponds to the contexts of the image such as speed and force.

We first introduce two labels representing timing and channels to represent the sub-sequences and channels that are introduced above.  
We then explain the model architecture using the labels, which is composed of an encoder, a generator, and a discriminator, and its training.
We finally show the method to predict an image at an arbitrary timing of a sequence from the beginning and ending images of the sequence.

\subsection{Label Presentation}

We introduce two labels: the timing label that represents the timing of a sequence and the channel label that represents the existence of each channel in the input images for the discriminator.   

\subsubsection{Timing label}

We consider the case that the sequence can be divided into a fixed number $N$ of sub-sequences or actions\footnote{We leave the dynamic estimation of number of sub-sequences for future work.}.
To represent timing in a continuous sub-sequence, we represent a timing label by the relative position of the timing in the sub-sequence. 
The timing label $\mathbf{t}_s$ of a step $s$ in a $n$-th sub-sequence ($0 < n$ ${\leq}$ $N$) with $S$ steps ($0$ ${\leq}$ $s$ ${\leq}$ $S$) is represented as the following $(N+1)$-dimensional vector:
\begin{equation}
\mathbf{t}_s=\left[0,..., \frac{s}{S}, \frac{S-s}{S}, ..., 0\right],
\end{equation}
where $n-1$-th and $n$-th dimensions has values that represent the relative position in the sub-sequence and other values are zero. 
This representation does not depend on the numbers of steps. The timing label of close steps has similar values, so the weights for the labels with close steps receives update information and this is expected to encourage the smooth changes in each sub-sequence. 

\subsubsection{Channel label}

A channel label is a label to denote which channels are included in the input to the discriminator. A channel is either a original channel, which corresponds to an image, or one of additional channels, which corresponds to a physical context.
This label consists of a set of binaries, each of which corresponds to a specific channel and becomes 1 when the input contains the channel, 0 otherwise. 

\begin{figure*}[t!]
\centering
\includegraphics[width=\linewidth]{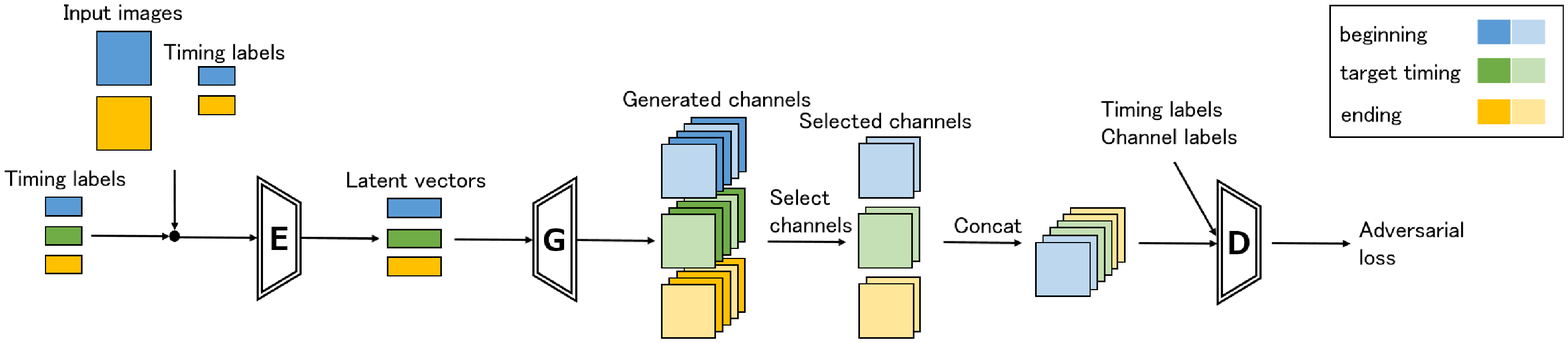}
\caption{\ourgan{} in training}
\label{figure:proposed_train}
\end{figure*}

\subsection{Model Architecture and Training}

In training, \ourgan{} consists of three modules: the encoder, the generator, and the discriminator. We show the overview of \ourgan{} in Figure~\ref{figure:proposed_train}.
The encoder (E) produces latent vectors from beginning and ending images, which have original image channels alone, their timing labels, and a timing label that denotes a target timing. The generator (G) then generates all the channels of the beginning, target timing, and ending from the latent vectors. 
The encoder and generator together generate all the channels of the beginning, target, and ending from two images and their labels and a target timing label.
Finally, the discriminator (D) receives specific channels of the generated beginning, target, and ending timing channels into real or not, using the timing labels and a channel label that specifies the target channels.
We explain the details of these modules and their training in the remaining of this section.

\subsubsection{Encoder}
The encoder receives beginning and ending images, their timing labels, and a target timing label, and encodes them into three latent vectors that corresponds to beginning, ending and target timings. 
The encoder consists of layered four blocks, each of which consists of 2-dimensional (2D) convolution, batch normalization, and the leaky ReLU activation function, and a final linear layer.

\subsubsection{Generator}
The generator receives each latent vector from the encoder and generates all the channels including an image channel and additional physical context channels as 2D matrices.  
The generator is composed of layered four blocks, each of which consists of 2D transposed convolution, batch normalization, and the leaky ReLU activation function, a final block with 2D transposed convolution with the $\tanh$ activation function. 

\subsubsection{Discriminator}
The discriminator receives real or generated $N_d$ channels in beginning, ending, and target timings, the timing labels, and the channel label specifying the selected channels and discriminates them into real or not. Here, $N_d$ is an integer to specify the number of channels for input of the discriminator and the $N_d$ channels are randomly selected from both original and additional channels.

The discriminator is composed of layered four blocks, each of which is a 2D spectral normalized convolution~\cite{SNGAN} with the leaky ReLU activation function, a global sum-pooling, a 2D spectral normalized linear layer, and a sigmoid activation function. 
Concat or Projection in Figure~\ref{figure:concat} are used for incorporating the labels. 
As for Projection, We set the first four blocks and the global sum-pooling to $\phi$, the 2D spectral normalized linear to $\phi'$, and the sigmoid function to $\phi''$.

\subsubsection{End-to-end training}

\begin{algorithm*}[t!]
\caption{PCTGAN using selected channel. We use $n_{disc}=5$, $n_{step}=3$, $n_{ch}=5$, $m=128$, $\beta_{1}=0.5, \beta_{2}=0.9$.}
\label{alg:pctgan}
\begin{algorithmic}[1]
\renewcommand{\algorithmicrequire}{\textbf{Input:}}
\renewcommand{\algorithmicensure}{\textbf{Output:}}
\REQUIRE The gradient penalty coefficient $\lambda$, the number of discriminator iterations per generator iteration $n_{disc}$, the number of steps per generator iteration $n_{step}$, the number of all channels $n_{ch}$, the number of selected channels $n_{s\_ch}$, the batch size $m$, Adam hyper-parameters $\alpha, \beta_{1}, \beta_{2}$, encoder $E_{\phi}$, generator $G_{\theta}$, discriminator $D_{w}$, initial encoder parameters $\phi_{0}$, initial generator parameters $\theta_{0}$, initial discriminator parameters $w_{0}$.
\ENSURE Encoder parameters $\phi$, generator parameters $\theta$, discriminator parameters $w$.
\STATE $\mathbb{A}_{ch} = \{1,...,n_{ch} \}$
\STATE $w = w_{0}$, $\theta=\theta_{0}$, $\phi=\phi_{0}$
\WHILE {$\theta$ has not converged}
    \FOR {$t = 1,...,n_{disc}$}
        \FOR {$i = 1,...,m$}
            \STATE Randomly select $n_{s\_ch}$ channels $\mathbb{A}_{s\_ch} \sim \mathbb{A}_{ch}$.
            \STATE Generating a channel label $l_{ch}$ from $\mathbb{A}_{s\_ch}$.
            \FOR {$k = 1,...,n_{step}$}
                \STATE Sample real images $x$ and labels $l_{time}$ in a $N_{k}$-step sequence $x^{(N_{k})}, l_{time}^{(N_{k})} \sim \mathbb{P}_{r}$.
                \STATE Select the original channel $x^{(N_{k})}_{1}$.
            \ENDFOR
            \FOR {$k = 1,...,n_{step}$}
                \STATE $\acute{z}^{(N_{k})} \leftarrow E_{\phi} \left( x^{(N_{1})}_{1}, x^{(N_{n_{step})}}_{1}, l_{time}^{(N_{1})}, l_{time}^{(N_{n_{step}})}, l_{time}^{(N_{k})} \right)$
                \STATE $\acute{x}^{(N_{k})} \leftarrow G_{\theta} \left( \tilde{z}^{(N_{k})} \right)$
                \STATE Sample a selected channel $\acute{x}^{(N_{k})} \leftarrow \{ \acute{x}^{(N_{k})}_{n} | n \in \mathbb{A}_{s\_ch} \}$.
            \ENDFOR
            \FOR {$k = 1,...,n_{step}$}
                \STATE Sample a selected channel $x^{(N_{k})} \leftarrow \{ x^{(N_{k})}_{n} | n \in \mathbb{A}_{s\_ch} \}$.
            \ENDFOR
            \STATE Concatenate all step images $\tilde{x} \leftarrow \{ \acute{x}^{(N_{k})} | k \in \{1,...,n_{step} \} \}$, $x \leftarrow \{ x^{(N_{k})} | k \in \{1,...,n_{step} \} \}$.
            \STATE $\epsilon \sim \mathbb{U}(0,1)$
            \STATE $\hat{x} \leftarrow \epsilon x + (1-\epsilon) \tilde{x}$
            \STATE $L^{(i)} \leftarrow D_{w}(\tilde{x},l_{time},l_{ch}) - D_{w}(x,l_{time},l_{ch}) + \lambda (\parallel \nabla_{\hat{x}} D_{w}(\hat{x},l_{time},l_{ch}) \parallel_{2} - 1)^2$
        \ENDFOR
        \STATE $w \leftarrow Adam(\nabla_{w}\frac{1}{m}\sum^{m}_{i=1} L^{(i)}, w, \alpha, \beta_{1}, \beta_{2})$
    \ENDFOR
    \STATE Sampling a batch of images $\{ \tilde{x}^{(i)} \}^{m}_{i=1}$.
    \STATE $\phi \leftarrow Adam(\nabla_{\phi}\frac{1}{m}\sum^{m}_{i=1}-D_{w}(\tilde{x},l_{time},l_{ch}), \phi, \alpha, \beta_{1}, \beta_{2})$
    \STATE $\theta \leftarrow Adam(\nabla_{\theta}\frac{1}{m}\sum^{m}_{i=1}-D_{w}(\tilde{x},l_{time},l_{ch}), \theta, \alpha, \beta_{1}, \beta_{2})$
\ENDWHILE
\end{algorithmic}
\end{algorithm*}

We train the model in an end-to-end manner. 
We build training instances, each of which is composed of three images randomly chosen from the training sequence. For the three training images of each training instance, the encoder receives the first and last training images as the beginning and ending images, the generator generate intermediate channels from the encoder's output, and the discriminator discriminates the training channels, which are the remaining intermediate training image and the physical contexts, with their timing and channel labels and the generated channels with their labels. We use the  loss function of WGAN-GP, which was shown in Eq.~(\ref{eq:WGAN-GP}).

We show the pseudo-code of our \ourgan{} using selected channel in Algorithm~\ref{alg:pctgan}.
We use $\phi$, $\theta$, $w$ for the parameters of encoder $E_{\phi}$, generator $G_{\theta}$ and discriminator $D_{w}$, respectively.
We train the entire model by repeating the update of the parameters so that theta converges, in other words, the discriminator loss $\leftarrow D_{w}(\tilde{x},l_{time},l_{ch})-D_{w}(x,l_{time},l_{ch})$ converges using the Adam optimizer.
We use $n_{ch}$ for the number of all channels, $\mathbb{A}_{ch}$ for a set of natural numbers ${1,...,n_{ch}}$, $n_{s\_ch}$ for the number of selected channels, and $\mathbb{A}_{s\_ch}$ for $n_{s\_ch}$ sampled randomly from $A_{ch}$ (line 6). 
$l_{ch}$ is a multi-class one-hot label, where the value of the dimension that corresponds to the channel number in $\mathbb{A}_{s\_ch}$ is 1 (line 7).
Images $x$ and timing labels $l_{time}$ are sampled from real data $\mathbb{P}_{r}$ in $n_{step}$ times (line 9).
The original channel image and labels $l_{time}$ are passed through the encoder and generator to obtain the all channels, that is, the original and additional channels (lines 12-16).
$\tilde{x}$ is made by concatenating selected channels $ch_{s\_ch}$ that are extracted from the obtained channels using $l_{ch}$ for all steps (lines 17-20).
The sum of the discriminator loss and gradient penalty $L^{(i)}$ are calculated, where $\hat{x}$ is linear sum of $\tilde{x}$ and $x$ using $\epsilon$ sampled from $\mathbb{U}(0,1)$ (lines 21-23).
After that, based on the discriminator losses, the parameters of encoder $E_{\phi}$, generator $G_{\theta}$, and discriminator $D_{w}$ are updated (lines 25,28,29).


\subsection{Prediction}

\begin{figure*}[t!]
\centering
\includegraphics[width=.7\linewidth]{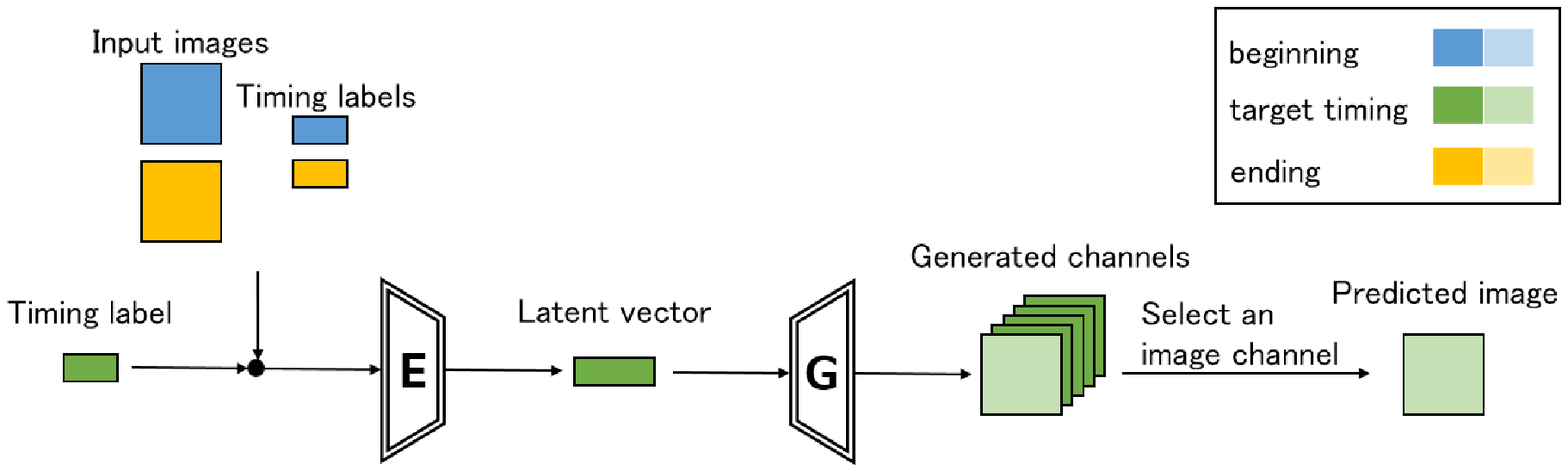}
\caption{\ourgan{} in prediction}
\label{figure:proposed_predict}
\end{figure*}

We generate an intermediate image at an arbitrary timing from the beginning and ending images in prediction using the encoder and generator as in Figure~\ref{figure:proposed_predict}. We use the original channel of a target timing as a predicted intermediate image.

\section{Task: Die Forging Process}

\begin{figure}[t!]
\centering
\includegraphics[width=\linewidth]{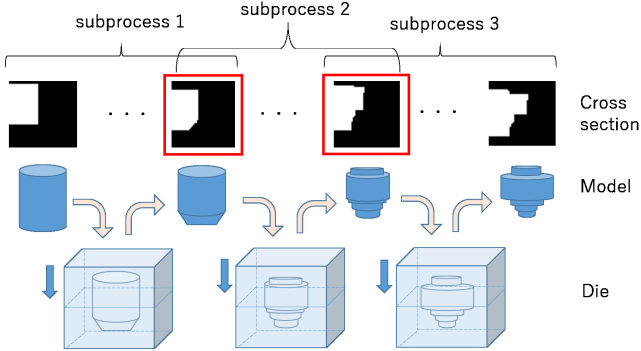}
\caption{An example die forging process. We assume the product models and dies are all rotationally symmetric. We use the shapes of the cross sections and represent the shapes as the white regions of binary images. The red boxes show the target intermediate die shapes in the designing task.}
\label{figure:molding_process}
\end{figure}

We target die forging processes of manufacturing, where metal is passed through several dies to be compressed to the target piece that matches the shape of the final die as shown in Figure~\ref{figure:molding_process}. Designing appropriate shapes of intermediate dies is important to produce strong and tough pieces. The dies are usually designed by repeating manual trial and error; expert designers design die candidates from the dies of existing pieces, which are similar to a target piece, and evaluate the candidates using design supporting tools that are based on high precision, time-consuming simulations, e.g., DEFORM\texttrademark~\cite{deform}. The computer-aided design is demanded to reduce the dependence on experts and to reduce the number of evaluations with simulations.

We define the designing task as follows. We assume each die forging process (sequence) consists of $N$ sub-processes, which corresponds to sub-sequences.\footnote{We use $N=3$ in this paper, but $N$ can be varied depending on the target task.} Each sub-process is further divided into steps, which corresponds to timings. Each step has its corresponding shape image. Figure~\ref{figure:molding_process} illustrates an example die forging process with $N=3$.
The shape in the first step of the first sub-process is the initial shape, and the shape in the final step of the last sub-process is the final shape, that is, the target shape. Note that these shapes are continuous between adjacent sub-processes and the first shape of a sub-process is the same as the last shape of the previous sub-process. The goal of the task is to generate the intermediate shapes shared by two sub-processes, i.e., the initial shapes of 2nd to $N$-th sub-processes, which are the same as the final shapes of 1st to ($N-1$)-th sub-processes. 

As in Figure~\ref{figure:molding_process}, we assume that the models and dies are all rotationally symmetric. We use the binary, black-and-white images of the shapes of the cross sections since we can reconstruct the models from the shapes by rotating predicted shapes.   

This task is different from other image interpolation tasks in two ways. First, we assume all the images and corresponding physical contexts in several processes are available in training. Other major datasets like CIFAR-10 and CelebA do not contain such sequence and additional physical information. 
Second, we assume that the processes have sub-processes. The sub-processes are usually not considered in other interpolation tasks. The task of estimating the changes of shapes is thus not a simple interpolation task. In this task setting, the changes in images are smooth in each sub-process but they are not smooth between sub-processes. 

\section{Experiments}

\subsection{Experimental Settings}
\subsubsection{Data preparation}

We prepared the images from the CAE dataset of die forging process. 

We map the shape information in the original CAE dataset into a 64x64 image as an original image channel, with setting pixels to 255 if the shapes exist and 0 otherwise.
When putting the images into \ourgan{}, we normalized the pixels into [-1, 1]. 

Similarly, we create the inputs of additional channels from four physical information, radial stress, axial stress, radial strain, and axial strain in the original CAE dataset.
We kept 99.7\% of values and clipped the remaining 0.3\% to ignore outliers and normalized the values into [-1, 1].
We then mapped the four physical information to 64$\times$64 matrices. When multiple values are mapped to a pixel, we chose the one with the largest absolute value as the pixel value.

The data set consists of 14 target product models with 12 modifications for each model. Each process has 3 sub-processes. 
The modifications include extensions and shrinkage, so all the target shapes are different among processes. 
The changes of shapes are generated with simulations, and they are manually checked so that the changes have no collapse. 
Each forging process is divided into about 80 time steps in total. Note that the numbers of steps depend on the process and they are not fixed. 
Among 14 models, 12 models with 14,102 shapes are used for training data set, another model with 1,150 shapes is used for development data set, and the remaining model with 938 shapes is used for test data set. 
In addition to the data set, we automatically generate training, development, and test data sets for the encoder using the generator. Each data set contains triples of shape and view images, latent vectors, and labels. We prepare 38,400 training, 3,200 development, and 3,200 test triples. 

\subsubsection{Training settings}

We train the model using Adam with ${\beta}_1=0.5$ and ${\beta}_2 = 0.9$~\cite{adam} with a mini-batch size of 128. 
For all the convolutions in the modules, we set the kernel size to [4, 4], stride to [2, 2] and padding to [1, 1].
We set the numbers of output channels of the block to [64, 128, 256, 64] for the convolution in the encoder, [256, 256, 128, 64] for the transposed convolution in the generator, and [64, 128, 256, 256] for the convolution in the discriminator. 
Other hyper-parameters are tuned with the successive halving pruner in a hyper-parameter optimization framework Optuna~\cite{optuna}. Such hyper-parameters include the learning rates of the encoder and generator, the learning rate of the discriminator using two time-scale update rule (TTUR)~\cite{TTUR}, the gradient penalty coefficient in WGAN-GP, the dimensions of latent vectors from the encoder, $\lambda$ in Eq.~(\ref{eq:WGAN-GP}) and the slope parameters of leaky ReLU activation functions in the modules. 

\subsubsection{Evaluation metric}

We evaluate the generated image channel using Fr\'echet Video Distance (FVD)~\cite{FVD}. FVD is a metric to measure the closeness between real and generated sequences by comparing the real images and generated images in a sequence level. 
We first extract features from real and generated images using Inflated 3D Convnet~\cite{I3D} pre-trained on action recognition task data sets UCF-101~\cite{Ucf101} and HMDB-51~\cite{Hmdb51} and calculate FVD using the mean and variance of the features as follows:
\begin{equation}
d( P_{r},P_{g})
= | \mu_{r} - \mu_{g} | + \mathrm{Tr}( \Sigma_{r} + \Sigma_{g} - 2 ( \Sigma_{r} \Sigma_{g} )^{\frac{1}{2}} ),
\end{equation}
where $P_{*}$, $\mu_{*}$, and $\Sigma_{*}$ are the distribution, mean, and co-variance matrix of images, respectively, and $r$ and $g$ denotes real and generated images.

\subsection{Results}

We evaluate the effects of timing labels and additional channels. We also investigate the number of input channels $N_d$ in the discriminator and compare the generated images.

\subsubsection{Effect of timing and physical contexts}\label{sec:with additional channel}
\begin{table}[t!]
 \begin{center}
  \scalebox{0.90}[0.9]{
  \begin{tabular}{lrrrr}
  \hline
  & \multicolumn{2}{l}{w/o additional channels} &  \multicolumn{2}{l}{w/ additional channels}\\
   &validation &test &validation &test\\
  \hline
  None &91.23 &91.36&63.74 &77.83\\
  Concat &47.08 &57.25 &47.02 &54.81\\
  Projection &59.96 &87.59 &33.82 &40.17\\
  \hline
  \end{tabular}}
  \caption{Effects of timing labels and additional channels in FVD. }
 \label{table:no_addtional_channel}
 \end{center}
\end{table}
To evaluate the use of timing labels, we trained our model using timing labels in three different ways: None, Concat, and Projection. None denotes training without timing label.
As for the evaluation of additional channels, we train our model only with generating original channel and with both generating original and additional channels.

We show the results in Table~\ref{table:no_addtional_channel}. 
With the same additional channels, Concat and Projection show better performance than None. This shows the generator can capture sequences with timing labels. 
As for the use of the additional channels, FVD consistently decreased by incorporating additional channels. 
Here, we set $N_d$ to 2 in the discriminator because FVD increased with more channels. We will discuss the effect of the number of channels in the next section.

\subsubsection{Analysis on the number of channels in discriminator}\label{sec:channel select}
\begin{table}[t!]
 \begin{center}
  \scalebox{0.90}[0.9]{
  \begin{tabular}{lrr}
  \hline
  $N_d$ &validation &test\\
  \hline
  1 &189.20 &183.31\\
  2 &33.82 &40.17\\
  3 &53.34 &67.36\\
  4 &47.90 &55.45\\
  5~(all) &49.90 &48.21\\
  \hline
  \end{tabular}}
 \end{center}
 \caption{FVD with the number of selected channels}
 \label{table:addtional_channel_chlabel}
\end{table}

We varied the number of input channels $N_d$ in the discriminator. We chose Projection for the use of timing label. 
The results are shown in Table~\ref{table:addtional_channel_chlabel}. 
The best FVD was achieved when $N_d = 2$.  
The high FVD with $N_d=1$ is reasonable because correspondence between channels is not explicitly shown to the discriminator.
The drop in the higher $N_d$ might be because it was difficult to model correspondences between channels.

\subsubsection{Visual comparison} 
\begin{figure}[t!]
\centering
\includegraphics[width=0.70\linewidth]{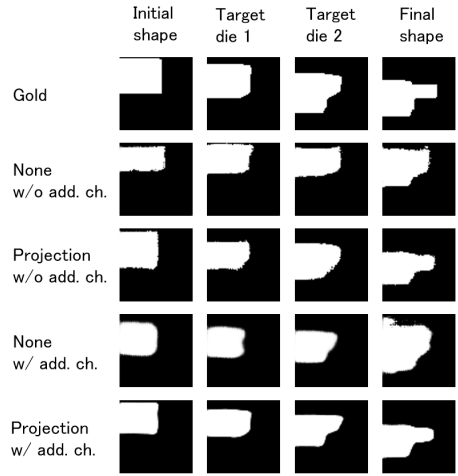}
\caption{Examples of generated images. }
\label{figure:result_shape}
\end{figure}
Figure~\ref{figure:result_shape} shows the generated examples of initial shape, intermediate target die shapes, and final shape by using four different models and their corresponding gold shapes.
We compare the models by disabling and enabling the shape and timing labels. We use Projection when we use the timing labels.
With both timing and additional channels, the model generated the closest sequence to the gold sequence.
The exterior shapes and positions are close to those of gold with the timing labels. This indicates the timing label holds the exterior shape and position information of the shapes. 
For additional channels, the edges of shapes with additional channels are smoother than those without additional channels. This shows that the additional channels are useful to make clear the outline of the shapes.
These results demonstrate the importance of simultaneously using both timing and physical information.

\section{Conclusions}

We proposed a novel method to generate images between two input images in a sequence. We introduced a novel \ourgan{} that models the sequence of images and the physical contexts behind the images using timing and channel labels. 
In the experimental evaluations to generate the shapes of dies in forging processes, we showed both the timing and physical information are effective in generating a sequence of images. 

For future work, we will investigate the better representation of timing, the use of explicit physical constraints in the model, and the application to other tasks. We also plan to evaluate the results in the manual design process of the dies.


\bibliographystyle{named}
\bibliography{ijcai20}

\end{document}